# Fuzzy C-Means-based Scenario Bundling for Stochastic Service Network Design

Xiaoping Jiang
School of Computer Science
University of Nottingham Ningbo China
Ningbo, China
Xiaoping.Jiang@nottingham.edu.cn

Ruibin Bai
School of Computer Science
University of Nottingham Ningbo China
Ningbo, China
ruibin.bai@nottingham.edu.cn

Dario Landa-Silva
School of Computer Science
University of Nottingham
Nottingham, UK
dario.landasilva@nottingham.ac.uk

Uwe Aickelin
School of Computer Science
University of Nottingham Ningbo China
Ningbo, China
uwe.aickelin@nottingham.edu.cn

*Abstract*—Stochastic service network designs with uncertain demand represented by a set of scenarios can be modelled as a large-scale two-stage stochastic mixed-integer program (SMIP). The progressive hedging algorithm (PHA) is a decomposition method for solving the resulting SMIP. The computational performance of the PHA can be greatly enhanced by decomposing according to scenario bundles instead of individual scenarios. At the heart of bundle-based decomposition is the method for grouping the scenarios into bundles. In this paper, we present a fuzzy c-means-based scenario bundling method to address this problem. Rather than full membership of a bundle, which is typically the case in existing scenario bundling strategies such as k-means, a scenario has partial membership in each of the bundles and can be assigned to more than one bundle in our method. Since the multiple bundle membership of a scenario induces overlap between the bundles, we empirically investigate whether and how the amount of overlap controlled by a fuzzy exponent would affect the performance of the PHA. Experimental results for a less-than-truckload transportation network optimization problem show that the number of iterations required by the PHA to achieve convergence reduces dramatically with large fuzzy exponents, whereas the computation time increases significantly. Experimental studies were conducted to find out a good fuzzy exponent to strike a trade-off between the solution quality and the computational time.

## I. Introduction

Service network design is used in less-than-truckload (LTL) transportation to address the selection, routing and scheduling of services, with the aim of making good profits while rendering excellent service [1,2]. As a matter of fact, the decision making in LTL transportation is subject to various uncertainties [3], amongst which the demand uncertainty is of particular importance. Two-stage stochastic programming is an appropriate framework for modeling service network design with uncertain demand, which is also referred to as stochastic service network design [3].

In most real-life applications, the uncertain demand is expressed in continuous probability distributions or discrete distributions with large numbers of outcomes [4]. Except for some trivial cases, two-stage stochastic programs with such inputs cannot be solved directly. The most widely applied approach to this problem consists of approximating these distributions as a limited number of discrete outcomes (scenarios), each with a known probability of occurrence. As the uncertain demand unfolds over time, these outcomes are organized into a hierarchical tree-like structure, termed the scenarios tree.

The two-stage stochastic program based on scenario trees can be formulated into what is referred to as the extensive form [5] or deterministic equivalent [6], which is essentially a large-scale deterministic model. Since directly solving the resulting model within acceptable computing time is typically beyond the capability of existing commercial solvers, various decomposition procedures have been developed to divide the extensive form into smaller, more manageable subproblems. As a scenario-based decomposition method, the progressive hedging algorithm (PHA) lends itself well to stochastic service network design where individual scenario problems can be solved efficiently. After decomposing the extensive form by scenarios into single-scenario subproblems, the PHA iteratively solves each subproblem and aggregates solutions of these subproblems into an overall solution that is implementable. This continues until a consensus solution amongst all of the subproblems is obtained or other stopping criteria such as time limit are met. Although the desirable theoretical property of convergence to a global optimum in the convex case [7] does not hold in the context of stochastic service network design, the PHA can be effectively used as a heuristic to find high-quality solutions within reasonable time [8].

The performance of the PHA can be enhanced by replacing the preceding scenario decomposition with bundle decomposition [5, 9], where individual scenarios are combined into bundles and the extensive form is decomposed using scenario bundles into multi-scenario subproblems. In contrast to scenario decomposition, the bundle version of the PHA solves smaller numbers of larger subproblems [7]. The bundle decomposition outperforms scenario decomposition by significantly cutting down the number of subproblems and hence the number of iterations required to reach a consensus. At the same time, this decrease should be carefully weighed against the increased difficulty of addressing each subproblem.

At the heart of bundle decomposition is the method for scenario bundling. This paper presents a fuzzy c-means (FCM) based method to partition the scenario set into bundles, which is distinctly different from existing scenario bundling methods to our knowledge. The FCM based scenario bundling computes a fractional membership score to measure the degree to which a scenario belongs to a bundle. In other words, a scenario belongs to each bundle to some degree and hence can be assigned to several bundles concurrently. By contrast, previous scenario bundling methods assigns each scenario to exactly one bundle. Our empirical tests confirm the effectiveness of FCM based scenario bundling in the PHA.

The remainder of this paper is organized as follows. In Section II, we review related literature on scenario bundling. The two-stage model of stochastic service network design is given in Section III. In Section IV, we present the fuzzy c-means-based method to group scenarios into bundles. In Section V, we introduce the progressive hedging algorithm which is used as a heuristic to solve the two-stage model. Experimental results are reported in Section VI. In Section VII, we give our concluding remarks and some directions for further study.

## II. LITERATURE REVIEW

In recent years, scenario bundling has received growing attention because it has the potential to yield striking improvements in the computational performance of the scenario-wise decomposition algorithm. Escudero et al. applied scenario bundling to two-stage stochastic mixed 0-1 problems for obtaining strong lower bounds [10]. They partitioned the scenario set into mutually exclusive bundles, each having the same cardinality. Hence, the number of bundles was chosen to be a proper divisor of the cardinality of the scenario set. However, the allocation of scenarios was random since no criterion concerning which scenarios would be grouped into the same bundle was given. Gade et al. incorporated scenario bundling into the PHA for two-stage SMIPs and present a method to compute lower bounds for the purpose of assessing the quality of the solutions generated by the PHA [9]. In their work, all bundles are of the same size and there are no common scenarios between any two bundles. Their computational results indicated that bundling scenarios could lead to dramatic improvements in the quality of lower bounds. Ryan et al. formulated the problem of determining scenario bundles for two-stage stochastic mixed 0-1 programs as a mixed integer program (MIP), the objective function of which is to maximize the bound improvement [11]. The proposed MIP formulation is based on the assumption that the intersection of any two scenario bundles is an empty set and the size of individual bundles is far less than the total number of scenarios. Crainic et al. present a k-means-based scenario bundling strategy [5]. They chose commodity demand as well as the solution information about single-scenario subproblems as scenario features and calculated the Euclidean distance between different scenarios to measure their similarity. The k-means method is used to group similar scenarios together and assign each scenario to only one bundle. Furthermore, the authors propose a covering strategy, where each scenario is assigned to its two closest bundles in terms of Euclidean distance. The proposed strategies are integrated as a preprocessing step into the progressive hedging-based meta-heuristic to solve the stochastic service network design problem. Their studies empirically show that, compared with randomly bundling or without scenario bundling, the k-means-based bundling strategy consumes less computing time to produce solutions of higher quality. In particular, the covering strategy with commodity demand as scenario features yields the best performance amongst all of the proposed strategies. Besides these authors' works, scenario bundling is being vigorously studied under different names in other contexts, such as the multistage stochastic programs [12] and the chance-constrained programs [13].

To summarize, all of the existing scenario bundling strategies for two-stage stochastic programs assign each scenario to exactly one bundle, except for the covering strategy. Inspired by the covering strategy, we relax this constraint and allow a scenario to belong to multiple bundles. Unlike the covering strategy, each scenario is not restricted to appear in precisely two bundles in this paper. Instead, we group scenarios based on their fractional membership scores and hence different scenarios can belong to different numbers of bundles.

## III. PROBLEM STATEMENT

To present the model of stochastic service network design, we introduce the following notation. Let $N$ represent the set of terminals in LTL transportation. The transportation service is scheduled over a time horizon of $T$ periods, denoted by $\{0,1,\ldots,T-1\}$. Let $K$ be the commodity set. Each commodity $k \in K$ is characterized by its quantity, which is stochastic and will be explained later, its origin terminal $o(k)$ and the time period $\sigma(k)$ when it is available, as well as its destination terminal $s(k)$ and the delivery deadline $\tau(k)$. We assume that the transport movement between every pair of terminals takes one period and use $t^-$ to represent the departure time of a movement with arrival time $t$. Because $t \in \{0,1,\ldots,T-1\}$, it follows that

$$t^- = \begin{cases} t-1 & t \geq 1 \\ T-1 & \text{otherwise} \end{cases} \quad (1)$$

We can obtain the so-called space-time network by duplicating each terminal in every time period [14]. Every pair of nodes in different time periods is connected by an arc, representing the transport movement from one terminal at a certain time period to another terminal at the next period. We use a 3-tuple $(i,j,t)$ to denote an arc from terminal $i$ at time period $t^-$ to terminal $j$ at time period $t$. In particular, an arc connecting the identical terminals in different time periods is referred to as a holding arc, which represents the activities of holding vehicles at a terminal for some time. Each arc $(i,j,t)$ has an associated fixed cost $c_{ij}$ incurred by the transportation or holding service. Except for the holding arcs, each arc has a resource capacity denoted by $u$.

The decision-making process of stochastic service network design has a two-stage structure. In the first stage, LTL carriers make decisions whether an arc should be included in the service network. For an arc $(i,j,t)$, this decision is denoted by a binary variable $x_{ij}^t$. The corresponding vector **x** stands for such decisions on all of the arcs. After the first-stage decisions are made, a realization of the uncertain demand is observed. As mentioned in Section I, we have a limited number of scenarios $s$ for possible future demand, each with a probability of occurrence $p_s$. These probabilities are non-negative and sum to 1. The collection of these scenarios is denoted by $S$. The demand of commodity $k$ in scenario $s$ is represented by $d_s^k$ and the vector $\mathbf{d}^s$ denotes the demand of all types of commodities in that scenario. In the second stage, LTL carriers determine the flow of commodity based on the demand realization. We use the decision variable $y_{ijk}^{st}$ to represent the flow of commodity $k$ on arc $(i,j,t)$ in scenarios $s$. In addition, we assume that customer demand must be met. To deal with the situations where the demand exceeds the capacity of the service network, LTL carriers need the flexibility to outsource some of the orders to external suppliers. Let the decision variable $Z^s(k)$ denote the amount of outsourcing for commodity $k$ in scenarios $s$, whereas $\lambda$ stands for the cost of outsourcing one unit of the commodity.

Stochastic service network design can then be modelled as follows [3]:

**Stage 1:**

$$\min\left\{\sum_{i\in N}\sum_{j\in N}\sum_{t=0}^{T-1} c_{ij} x_{ij}^t + \lambda \sum_{s\in S} p_s Q_1(\mathbf{x},\mathbf{d}^s)\right\} \quad (2)$$

s.t.

$$\sum_{j\in N} x_{ji}^{t^-} = \sum_{j\in N} x_{ij}^t \quad \forall i, \forall t \quad (3)$$

$$x_{ij}^t \in \{0,1\} \quad \forall i, \forall j, \forall t \quad (4)$$

**Stage 2:**

$$Q_1(\mathbf{x},\mathbf{d}^s) = \min \sum_{k\in K} Z^s(k) \quad (5)$$

s.t.

$$\sum_{k\in K} y_{ijk}^{st} \leq u x_{ij}^t \quad \forall i, \forall j, \forall t, \forall i\neq j \quad (6)$$

$$-\sum_{j\in N} y_{jik}^{st^-} + \sum_{j\in N} y_{ijk}^{st} =$$

$$\begin{cases} d_s^k - Z^s(k) & \text{if } (i,t) \text{ is supply node for } k \\ -d_s^k + Z^s(k) & \text{if } (i,t) \text{ is demand node for } k \quad \forall i,\forall t,\forall k \\ 0 & \text{otherwise} \end{cases} \quad (7)$$

$$y_{ijk}^{s\tau(k)} = 0 \quad \forall i, \forall j, \forall k \quad (8)$$

$$y_{ijk}^{st} \geq 0 \quad \forall i, \forall j, \forall k, \forall t \quad (9)$$

$$Z^s(k) \geq 0 \quad \forall k \quad (10)$$

The objective function (2) minimizes the cost of constructing the service network, plus the expected outsourcing cost across all the scenarios. Constraint (3) is the design balance constraint, which ensures that the number of incoming and outgoing vehicles at each node is balanced. Constraint (4) enforces the binary restrictions on design variables. The objective function (5) minimizes the outsourcing cost for a scenario. Constraint (6) ensures that the total flow along each arc does not exceed its capacity. Constraint (7) makes sure that all the commodities are shipped from their origins to destinations and hence all customer demand is met. Constraint (8) ensures that a commodity must not flow past its delivery deadline. Constraints (9) and (10) are used to guarantee the non-negativity of decision variables.

## IV. FUZZY C-MEANS-BASED SCENARIO BUNDLING

For scenario-wise decomposition, the number of sub-problems increases linearly with the size of the scenario tree. The number of sub-problems can be significantly reduced by combining individual scenarios into bundles and decomposing the extensive form according to scenario bundles into multi-scenario subproblems, but at the cost of increasing the size of each subproblem. The benefits of scenario bundling usually far outweigh the increased difficulty of solving each subproblem, provided that the multi-scenario subproblems are manageable.

In many previous scenario bundling strategies, each scenario belongs to exactly one bundle and the set of scenarios are partitioned into several non-overlapping bundles. This form of partitioning is referred to as hard clustering in machine learning or cluster analysis. As mentioned in Section II, almost all of the existing scenario bundling strategies perform hard clustering. For instance, k-means-based scenario bundling assigns a bundle to each scenario and partitions the set of scenarios into mutually exclusive bundles, with the aim of grouping similar scenarios together into one bundle while separating different bundles as well as possible.

In k-means-based scenario bundling, we are often faced with the problem that many scenarios have some degree of uncertainty in their bundle membership. It is quite common to see some scenarios that lie in between the centers of several bundles. For such a scenario, each scenario has approximately the same degree of membership for these bundles. Although there is no clear preference of one bundle over the other in such cases, k-means-based scenario bundling assigns only one bundle to such scenarios. Thus it seems reasonable to relax the constraint about hard clustering and allow a scenario to belong to multiple bundles. Moreover, multiple bundle membership for a scenario could have a beneficial effect on the convergence rate for the PHA. If two bundles have a few scenarios in common, the difference between the solutions of the corresponding multi-scenario subproblems is expected to be smaller by comparison to the case of disjoint bundles in hard clustering. The smaller difference between the first-stage decisions will make the PHA easier to converge. In addition, the strategy of assigning each scenario to precisely two bundles has proved efficient in recent studies [5]. These potential benefits motivate us to eliminate the constraint about hard clustering and consider an alternative scenario bundling strategy that allows multiple bundle membership for a scenario.

Fuzzy c-means (FCM) is a data clustering algorithm that allows a data point to have membership in more than one cluster [15]. We present our method of scenario bundling based on FCM. In FCM-based scenario bundling, a scenario has partial membership in each of the bundles, rather than absolute membership in a single bundle as it is the case in hard clustering. More specifically, FCM-based scenario bundling computes a fractional membership score to measure the degree to which a scenario belongs to a bundle. As mentioned in Section III, the set of scenarios $s$ is denoted by $S$. We use $g$ to represent the number of bundles required and the collection of bundles $b$ is denoted by $B$ where $b \subseteq S$. The center of a bundle $b$ is represented by $v_b$ and the corresponding set of bundle centers is represented by $V$. The membership scores for all of the scenarios across all of the bundles can be organized into a matrix $\Delta$ called the fuzzy partition matrix, with the element $\delta_{sb}$ of row $s$ and column $b$ representing the degree of membership for the scenario $s$ in the bundle $b$. Each membership score is less than 1 and each column of the fuzzy partition matrix sums to 1, that is,

$$\delta_{sb} \in [0,1] \quad \forall s \in S \ ; \ \forall b \in B \tag{11}$$

$$\sum_{b \in B} \delta_{sb} = 1 \quad \forall s \in S \tag{12}$$

FCM-based scenario bundling decides the bundle centers and the membership scores by minimizing an objective function that is the sum of distance between a scenario and a bundle center weighted by that scenario's membership score over all possible scenario and bundle center pairs [16]. This objective function denoted by $J(\Delta, V)$ is given as follows.

$$J(\Delta, V) = \sum_{s \in S} \sum_{b \in B} (\delta_{sb})^m \| s - v_b \|^2 \tag{13}$$

Here $m (m > 1)$ is the exponent for the fuzzy partition matrix, which controls how much the resulting bundles can overlap with one another. A higher value of the exponent generally leads to a larger number of scenarios that have significant membership in multiple bundles.

The most common method for solving the minimization problem (13) is the alternating optimization algorithm [15], which is an iterative procedure. First, the fuzzy partition matrix is initialized by a random number generator. To satisfy constraint (12), each entry in the matrix is divided by the sum of the elements in the same column. Next, the bundle centers and the membership scores are sequentially updated and the objective function is calculated in each iteration. The iterative process continues until the maximum number of iterations specified is exceeded, or the improvement in the objective function between two consecutive iterations is less than a specified minimum threshold. The iterative procedure eventually outputs the final fuzzy partition matrix, with each scenario having a membership score for each bundle. We adopt the alternating optimization algorithm to calculate the bundle membership scores, which is shown in Phase 1 of Algorithm 1.

The resulting fuzzy partition matrix offers the flexibility to design various scenario bundling algorithms. For example, scenario bundling based on hard clustering can be implemented by assigning a scenario to the bundle with the highest membership score. In our proposed method, we define a membership score threshold $\gamma$ to determine whether to place a scenario into a certain bundle. A scenario will be assigned to those bundles in which the membership score of this scenario is greater than the specified threshold. This rule is meant for the cases where a scenario lies close to the center of a bundle and thus has a much higher membership score for this bundle than other bundles. For such scenarios, we have a clear preference of one bundle over the others. In other cases, a scenario lies in between the centers of the bundles and somewhat far away from the center of any bundle. Such a scenario will have a lower membership score for all the bundles, with the maximum score less than the given threshold $\gamma$. We denote the maximum score of a scenario $s$ by $h(s)$, where $h(s) = \max_{b \in B} \delta_{sb}$. To deal with these cases, we define an interval parameter $\eta \in [0,1]$. For a scenario whose maximum score $h(s)$ is less than the threshold $\gamma$, we assign it to those bundles in which the membership score goes above $\eta * h(s)$. Eventually, each scenario is placed into at least one bundle.

In our proposed method, a scenario may appear multiple times in the resulting bundles. At the same time, the sum of the probabilities over all the scenario bundles must be 1 regardless of the clustering method. Therefore, we cannot simply add together the probabilities of each scenario within a bundle to calculate the probability of this scenario bundle, as in the case of hard clustering. Instead, we first count the total number of times a scenario $s$ appears in all the final bundles. Then the original probability $p_s$ of this scenario is divided by its number of occurrences and we obtain the new probability of this scenario, denoted by $q_s$. The probability of a scenario bundle $p_b$ is obtained by summing the new probabilities of each scenario within this bundle. It is worth noting that our approach to calculating the probability of a scenario bundle equally applies to the case of hard clustering where the number of occurrences of each scenario is one.

The pseudocode for fuzzy c-means-based scenario bundling is given in Algorithm 1. Scenario bundling is accomplished through three phases. In the first phase, we calculate the bundle membership score using the alternating optimization method mentioned above. In the following phases, we combine scenarios into bundles and yield the probability of each bundle. In addition, we choose commodity demand as the scenario feature to measure scenario similarity. Other features, such as the solution-related features, are not considered in this paper because demand-based features produced higher quality solutions than solution-based features in previous scenario bundling strategies based on k-means [5].

**Algorithm 1: Scenario bundling based on fuzzy c-means**

**Input:** the set $S$ of scenarios $s$, the probability of each scenario $p_s$, the number of bundles $g$, the maximum number of iterations $\Lambda$, the minimum objective function improvement $\varepsilon$, the exponent for fuzzy partition matrix $m$, the membership score threshold $\gamma$, the interval parameter $\eta$.

**Output:** the set $B$ of bundles $b$, the probability of each bundle $p_b$.

**Begin**

**Phase 1: Calculate the bundle membership scores**

1: Initialize the membership score matrix $\Delta$ with uniformly generated random numbers between 0 and 1.
2: Normalize each entry in $\Delta$ through dividing it by the sum of the corresponding column.
3: **while** $LoopIndex < \Lambda$ and $|J_{new} - J_{old}| > \varepsilon$ **do**
4:    Calculate the bundle centers according to
$$\mathbf{v}_b = \sum_{s \in S}(\delta_{sb})^m s \bigg/ \sum_{s \in S}(\delta_{sb})^m, \ \forall b \in B$$
5:    Update the membership scores according to
$$\delta_{sb} = \left[ \sum_{l=1}^{g} \left( \frac{\|s - \mathbf{v}_b\|}{\|s - \mathbf{v}_l\|} \right)^{\frac{2}{m-1}} \right]^{-1} \begin{array}{l} \forall s \in S \\ \forall b \in B \end{array}$$
6:    Calculate the objective function $J$.
7: **end while**

**Phase 2: Place scenarios into bundles**

8: **for** each scenario $s \in S$ **do**
9:    **if** its maximum score $h(s) > \gamma$ **then**
10:       Assign $s$ to bundles whose scores are above $\gamma$.
11:    **else then**
12:       Assign $s$ to bundles whose scores fall within interval $[\eta * h(s), h(s)]$
13: **end for**

**Phase 3: Calculate the probability of each bundle**

14: Count the total number of times a scenario $s$ appears in all the final bundles.
15: Compute the new probability $q_s$ of $s$ through dividing its original probability $p_s$ by its number of occurrences.
16: Calculate the probability of a bundle $p_b$ according to
$$p_b = \sum_{s \in b} q_s, \ \forall b \in B$$

**End**

## V. PROGRESSIVE HEDGING HEURISTIC

### A. Bundle-based Decomposition

After obtaining the scenario bundles, we can decompose the two-stage model for stochastic service network design according to scenario bundles instead of individual scenarios. To do that, we first split each design variable $x_{ij}^t$ into several parts, one $x_{ij}^{tb}$ for each scenario bundle $b$, that is,

$$x_{ij}^t = \sum_{b \in B} p_b x_{ij}^{tb} \quad (14)$$

At the same time, the non-anticipativity constraints require that the design variables for different bundles should be identical, that is,

$$x_{ij}^{tb} = x_{ij}^{te}, \forall b, e \in B, b \neq e \quad (15)$$

The non-anticipativity constraints can be generally expressed as $AX = 0$, where $A$ is a coefficient matrix and $X$ is a vector of $x_{ij}^{tb}$.

Substituting (14) into (2) and simplifying the formula, we have the following objective function:

$$\min \sum_{b \in B} \left( \sum_{i \in N} \sum_{j \in N} \sum_{t=0}^{T-1} (c_{ij} p_b x_{ij}^{tb}) + \lambda \sum_{s \in b} \sum_{k \in K} (q_s Z^s(k)) \right) \quad (16)$$

The objective function can now be decomposed by scenario bundles and the newly added non-anticipativity constraints are the only ones that tie together different scenario bundles. The augmented Lagrangian method for equality constrained optimization can be used to address this problem [6]. The non-anticipativity constraints are added to the objective function via the Lagrangian multipliers $\alpha$ and a penalty term scaled by the penalty factor $\rho/2$. Thus we have

$$\min \left\{ \sum_{b \in B} \left( \sum_{i \in N} \sum_{j \in N} \sum_{t=0}^{T-1} (c_{ij} p_b x_{ij}^{tb}) + \lambda \sum_{s \in b} \sum_{k \in K} (q_s Z^s(k)) \right) + \alpha^T AX + \frac{1}{2} \rho \| X - \overline{X} \| \right\} \quad (17)$$

where $\overline{X}$ represents the expectation of the design variable $X$. For a given $\overline{X}$, both the objective function and constraints are now separable with respect to scenario bundles, hence the original problem decomposes by scenario bundles.

### B. Enforcing the Non-anticipativity Constraints

With bundle-based decomposition, the two-stage model for stochastic service network design is divided into a system of manageable multi-scenario sub-problems. The solutions to these sub-problems satisfy all of the constraints except for the non-anticipativity constraints. A sub-problem solution is said to be implementable if it fulfils the non-anticipativity constraints [17]. The PHA enforces the non-anticipativity constraints by iteratively aggregating the multi-scenario sub-problem solutions into an implementable solution until the sub-problem solutions agree with the implementable solution. Specifically, the PHA involves three key steps stated as follows.

**Step 1** Solve each multi-scenario sub-problem to obtain the solution to a first-stage design variable denoted by $\left(x_{ij}^{tb}\right)^{(r)}$.

**Step 2** Compute an implementable solution for this design variable according to

$$\left(\overline{x_{ij}^t}\right)^{(r)} \leftarrow \sum_{b \in B} p_b \left(x_{ij}^{tb}\right)^{(r)} \quad (18)$$

**Step 3** Update the dual variable associated with this design variable according to

$$\left(w_{ij}^{tb}\right)^{(r)} \leftarrow \left(w_{ij}^{tb}\right)^{(r-1)} + \rho\left(\left(x_{ij}^{tb}\right)^{(r)} - \overline{\left(x_{ij}^{t}\right)}^{(r)}\right) \quad (19)$$

The notations used here are explained. A variable with a parentheses-enclosed loop index $(r)$ in the superscript represents its value in the $r$ th iteration. $\overline{x_{ij}^{t}}$ is an element of $\overline{X}$ whereas the dual variable $w_{ij}^{tb}$ is an element of the coefficient vector of the term $\alpha^{T} AX$ in (17).

This process is repeated to generate a sequence of implementable solutions that provably converges to global optimum in the convex case [7]. However, the integrality constraints on the first-stage decision variables render our problem non-convex and convergence is not guaranteed. In such cases, the PHA is used as a heuristic method. On the other hand, the integrality constraints make it easier to judge whether two first-stage decisions are equal, which helps detect the convergence and define the stopping criteria. The PHA is deemed convergent if the difference between any two first-stage decisions is below a given tolerance threshold [8]. We also limit the maximum amount of time spent on iterating to hedge against the risk of non-convergence.

## VI. COMPUTATIONAL EXPERIMENTS

We evaluate the proposed scenario bundling method on a LTL transportation network adapted from Bai et al. [3], which consists of 12 nodes. Every pair of distinct nodes is connected by an arc. Each arc has a capacity of 12 units. The planning horizon considered in this experiment is divided into 5 periods and the transport movement between every pair of nodes takes one period. There are 6 commodities to be delivered. The outsourcing cost for one unit of the commodity is 80.

In this experiment, we utilize the triangular distribution Tri(5, 11, 8), which was also used by Lium et al. [4] and Bai et al. [3], to represent the demand uncertainty. Regarding the scenario generation, we employ the publicly available scenario generator from Høyland, Kaut and Wallace [18] to construct five scenario trees to approximate this distribution, each with a different number of scenarios. Unless otherwise indicated, all of the computing was conducted on a laptop computer with eight 2.80 GHz Intel Core i7 CPUs and 8 GB of RAM, under a 64-Bit operating system.

We implemented the proposed scenario bundling algorithm in MATLAB R2015b, making use of the pre-built function for fuzzy c-means clustering. The input arguments of this algorithm are specified in TABLE I. The interval parameter in the table defines the left endpoint of the interval by giving the percentage of the maximum score.

In order to investigate whether the amount of fuzzy overlap between the resulting bundles has any bearing on the performance of the following PHA, we chose nine different fuzzy partition matrix exponents $m$ for each scenario tree. The scenario bundling results are presented in TABLE II-IV. In the following, because of limited space, we take as an example the results for the scenario tree with 150 scenarios. According to TABLE I, this scenario tree is partitioned into seven bundles.

TABLE II shows the number of scenarios in each bundle. We observe that the sum of the number of scenarios over all the scenario bundles increases gradually with the fuzzy exponent when the exponent is less than 1.85. However, the number became much larger when the exponent goes above 1.85. The significant increase in the number of scenarios within a bundle would probably lead to much more computing time for solving the bundle subproblem.

TABLE III shows the number of scenarios that have been assigned to more than one bundle in each scenario tree, whereas the number of occurrences of each repeated scenario in the case of 150 scenarios with fuzzy exponent 2 is presented in TABLE IV.

TABLE I. Parameter settings for fuzzy c-means based scenario bundling

| #Sces | #Bundles $g$ | Score Threshold $\gamma$ | Interval Parameter $\eta$ |
|---|---|---|---|
| 48 | 5 | 0.8 | 0.95 |
| 72 | 7 | 0.8 | 0.95 |
| 96 | 7 | 0.8 | 0.95 |
| 120 | 7 | 0.8 | 0.95 |
| 150 | 7 | 0.8 | 0.95 |

TABLE II. The number of scenarios for each bundle in the case of 150 demand scenarios

| Bundle \ $m$ | 1 | 2 | 3 | 4 | 5 | 6 | 7 | Sum |
|---|---|---|---|---|---|---|---|---|
| 1.5 | 25 | 25 | 21 | 9 | 22 | 34 | 20 | 156 |
| 1.6 | 25 | 23 | 10 | 24 | 26 | 28 | 25 | 161 |
| 1.7 | 25 | 14 | 26 | 24 | 23 | 27 | 23 | 162 |
| 1.75 | 24 | 26 | 29 | 18 | 19 | 22 | 25 | 163 |
| 1.8 | 22 | 9 | 28 | 29 | 23 | 30 | 28 | 169 |
| 1.85 | 35 | 31 | 40 | 14 | 33 | 32 | 23 | 208 |
| 1.9 | 29 | 24 | 31 | 55 | 42 | 37 | 41 | 259 |
| 1.95 | 64 | 46 | 41 | 29 | 32 | 33 | 34 | 279 |
| 2 | 55 | 39 | 39 | 67 | 41 | 42 | 43 | 326 |

TABLE III. The number of repeated scenarios for each scenario tree under different fuzzy exponent settings

| #Sces \ $m$ | 48 | 72 | 96 | 120 | 150 |
|---|---|---|---|---|---|
| 1.5 | 1 | 2 | 1 | 6 | 5 |
| 1.6 | 2 | 2 | 7 | 5 | 10 |
| 1.7 | 5 | 5 | 7 | 9 | 11 |
| 1.75 | 6 | 7 | 9 | 10 | 11 |
| 1.8 | 7 | 7 | 14 | 12 | 16 |
| 1.85 | 15 | 12 | 13 | 44 | 42 |
| 1.9 | 12 | 21 | 23 | 52 | 60 |
| 1.95 | 36 | 34 | 19 | 59 | 51 |
| 2 | 47 | 40 | 38 | 92 | 57 |

TABLE IV. The number of occurrences of each repeated scenario in the case of 150 demand scenarios with exponent 2

| Sce | F | Sce | F | Sce | F | Sce | F | Sce | F | Sce | F |
|---|---|---|---|---|---|---|---|---|---|---|---|
| 3 | 4 | 19 | 5 | 61 | 5 | 79 | 3 | 106 | 4 | 138 | 4 |
| 4 | 5 | 23 | 3 | 62 | 2 | 84 | 4 | 107 | 2 | 140 | 2 |
| 5 | 6 | 25 | 2 | 64 | 2 | 85 | 6 | 108 | 2 | 141 | 6 |
| 7 | 6 | 28 | 5 | 65 | 4 | 88 | 5 | 112 | 3 | 144 | 5 |
| 9 | 4 | 30 | 4 | 67 | 4 | 92 | 7 | 118 | 5 | 146 | 4 |
| 10 | 2 | 33 | 5 | 68 | 6 | 95 | 6 | 119 | 6 | 147 | 2 |
| 11 | 3 | 37 | 5 | 70 | 3 | 98 | 6 | 127 | 4 | 148 | 5 |
| 13 | 2 | 38 | 5 | 72 | 5 | 102 | 7 | 128 | 6 | | |
| 15 | 3 | 52 | 2 | 76 | 6 | 104 | 3 | 130 | 2 | | |
| 16 | 2 | 53 | 5 | 77 | 2 | 105 | 4 | 131 | 3 | | |

As an overall trend, we observe that larger values of the fuzzy partition matrix exponent indicate a greater degree of fuzzy overlap between the resulting bundles. Specifically, the number of repeated scenarios under different exponent settings roughly follows the exponential growth of each scenario tree, where the numbers do not grow rapidly until the exponent reaches around 1.85. The average number of occurrences of repeated scenarios generally keeps an upward tendency with the increase of the fuzzy exponent.

We implemented the bundle version of the PHA in the C++ programming language in Microsoft Visual Studio 2010. Regarding the stopping criteria for the PHA, we took $10^{-5}$ as the tolerance threshold to measure the difference between any two first-stage decisions and 3 hours as the maximum computational time allowed. To deal with possible endless loops amid the iterative process of the PHA, we tested a set of different values {0.8, 1, 1.3, 1.5, 1.7, 1.9, 2} for the penalty factor and chose the one producing the best performance. For a scenario bundle, its associated network design model was created with CPLEX Concert Technology and solved by the CPLEX Mixed Integer Programming Optimizer in version 12.6.2. Except for a relative MIP gap tolerance of 0.05, all of the parameters controlling the behavior of CPLEX assume their default values. For the sake of comparison, the scenarios in each scenario tree are also partitioned by the k-means method into the same number of bundles as in the fuzzy c-means method and the corresponding network design models are solved in the same way. To start the iterative procedure of k-means, we select the initial bundle centroid positions by means of the k-means++ method [19]. As for the distance measure, we used the squared Euclidean distance. The network design results obtained using different scenario bundling strategies are summarized in TABLE V, where the computational time are rounded to the nearest integers. Here we take the objective function value obtained by k-means as the reference value and show the relative difference between the objective value yielded by FCM and the corresponding reference value, which is calculated according to

$$\frac{Obj - Obj_{reference}}{Obj_{reference}} * 100\% \qquad (20)$$

For each scenario tree, we observe that the objective function values were obtained for all of the fuzzy exponent settings. In other words, the bundle version of the PHA heuristic succeeds in finding feasible solutions to the original problem under different fuzzy exponent settings. This observation suggests that the bundle version of the PHA heuristic remains effective in the case of soft clustering, where a scenario can be assigned to more than one bundle. We look further into the solution quality by comparing the objective values from the two methods. It can be found that 80% of solutions obtained by FCM are of equal or better quality than those found by k-means. Amongst the inferior solutions yielded by FCM, the worst one has a relative difference of 2.1%.

For each scenario tree, we also observe that the number of iterations required to achieve convergence reduces dramatically when the fuzzy exponent is greater than 1.85. By contrast, the k-means based method needs roughly twice that number. The significant decrease brought about by larger exponents accords with our prior observation that the exponents above 1.85 lead to a great degree of fuzzy overlap between the resulting bundles. As a high degree of overlap indicates little difference between bundles, the subproblem solutions across different bundles are expected to share a lot of similarities and few iterations are needed to reach consensus among all of the subproblem solutions. In such cases, the computing time, however, nearly doubled in comparison with k-means. This is due to the fact that a higher degree of overlap typically results in a greater size of the subproblem for each bundle, thus increasing the difficulty of solving each subproblem. When the fuzzy exponent goes beyond 1.85, the additional time spent on each subproblem far outweighs the reduction in computational time resulting from fewer iterations.

For each scenario tree, it can be seen that both the number of iterations and computing time fluctuate significantly when the fuzzy exponent rise from 1.5 to 1.85. Under this condition, there are 15 solutions consuming less computational time than for k-means, which accounts for 50% of the total. Among the different sets of scenario bundles produced by FCM, we can find for each scenario tree at least one set that is better than the set produced by k-means. For instance, the set of scenario bundles produced in the case of 150 scenarios with exponent 1.6 makes the PHA heuristic consume 29.09% less time than the set produced by k-means. These results demonstrate that the fuzzy c-means based scenario bundling has the potential to greatly speed up the PHA heuristic compared to K-means when the parameter settings are appropriate.

TABLE V. Network design results obtained using different scenario bundling strategies

| #Sces | 48 | | | 72 | | | 96 | | | 120 | | | 150 | | |
|---|---|---|---|---|---|---|---|---|---|---|---|---|---|---|---|
| m | *Obj* | *#Ite* | *T(s)* | *Obj* | *#Ite* | *T(s)* | *Obj* | *#Ite* | *T(s)* | *Obj* | *#Ite* | *T(s)* | *Obj* | *#Ite* | *T(s)* |
| K-means | 136 | 8 | 252 | 140 | 12 | 359 | 144 | 11 | 385 | 140 | 9 | 691 | 137 | 13 | 845 |
| 1.5 | 0% | 9 | 221 | +0.7% | 17 | 446 | 0% | 12 | 434 | 0% | 27 | 652 | 0% | 20 | 818 |
| 1.6 | 0% | 15 | 328 | 0% | 15 | 337 | 0% | 11 | 472 | +0.7% | 7 | 490 | 0% | 13 | 739 |
| 1.7 | +0.7% | 24 | 405 | 0% | 13 | 364 | +0.7% | 21 | 629 | 0% | 12 | 559 | 0% | 11 | 782 |
| 1.75 | +0.7% | 7 | 181 | 0% | 13 | 414 | -2.1% | 17 | 485 | -1.4% | 18 | 676 | +0.7% | 13 | 841 |
| 1.8 | +1.5% | 13 | 221 | 0% | 11 | 395 | -0.7% | 10 | 380 | +0.7% | 8 | 621 | 0% | 22 | 962 |
| 1.85 | 0% | 24 | 625 | 0% | 6 | 435 | -2.8% | 13 | 536 | 0% | 14 | 1026 | 0% | 11 | 748 |
| 1.9 | 0% | 4 | 181 | 0% | 5 | 707 | 0% | 9 | 423 | 0% | 9 | 2035 | 0% | 9 | 1403 |
| 1.95 | 0% | 8 | 407 | 0% | 7 | 859 | 0% | 13 | 550 | 0% | 4 | 1362 | 0% | 8 | 1438 |
| 2 | 0% | 4 | 586 | +2.1% | 9 | 918 | -2.8% | 12 | 977 | 0% | 6 | 2610 | 0% | 6 | 1200 |

## VII. Conclusions And Future Directions

In this paper, we present an alternative scenario bundling method based on FCM. The proposed method differs from existing scenario bundling methods in that a scenario is allowed to appear in more than one bundle. Therefore, our method lends itself well to dealing with the common cases where a scenario has approximately the same degree of membership in several bundles. The multiple bundle membership for a scenario induces some overlap between the resulting bundles. The amount of overlap can be controlled by adjusting the value of the fuzzy partition matrix exponent. When the exponent is greater than 1.85, the number of iterations required by the PHA to achieve convergence reduces dramatically. However, the computation time increases significantly. When the exponent is less than 1.85, on our test instance the proposed method is competitive with, and often superior to, the k-means-based scenario bundling method.

In the future, the proposed method can be further explored in several directions. As the PHA is used as a heuristic in our problem, the lower bound information is important to assess the quality of the solutions. Recent studies suggest that the quality of the lower bounds obtained from the PHA can be improved dramatically by combining scenarios into disjoint bundles. Thus, it appears interesting to investigate whether and how FCM-based scenario bundling would enhance the quality of the lower bounds. Although we studied the impact of overlap in the context of two-stage stochastic service network design, the benefits of having overlap between scenario bundles naturally apply to the multi-stage SMIPs. Therefore, we will consider extending FCM-based scenario bundling method to the multi-stage SMIPs.

## Acknowledgment

This work is funded by the National Natural Science Foundation of China [grant number NSFC 71471092]; Zhejiang Natural Science Foundation [grant number LR17G010001]; Ningbo Science and Technology Bureau [grant number 2011B81006], [grant number 2014A35006]. We also greatly acknowledge the support from the International Doctoral Innovation Centre (IDIC) scholarship scheme.